\documentclass[sigconf]{acmart}
\usepackage{booktabs}
\usepackage{pifont}
\usepackage{wrapfig}
\usepackage{multirow}
\usepackage{wrapfig}
\usepackage{xspace}
\usepackage{amsmath}
\usepackage{float}
\usepackage{booktabs}
\usepackage{pifont}
\usepackage{wrapfig}
\usepackage{multirow}
\usepackage{wrapfig}
\usepackage{xspace}
\usepackage{amsmath}
\usepackage{float}
\usepackage{paralist}
\usepackage{enumitem}

\AtBeginDocument{%
  \providecommand\BibTeX{{%
    \normalfont B\kern-0.5em{\scshape i\kern-0.25em b}\kern-0.8em\TeX}}}

\renewcommand\footnotetextcopyrightpermission[1]{} 
\begin{document}
\pagestyle{empty} 
\title{GauU-Scene V2: Assessing the Reliability of Image-Based
Metrics with Expansive Lidar Image Dataset Using 3DGS and
NeRF}

\author{Butian Xiong }
\affiliation{%
  \institution{The Chinese University of Hong Kong, Shenzhen}
  \country{Mainland, China}
  }
\email{butianxiong@link.cuhk.edu.cn}

\author{Nanjun Zheng}
\affiliation{%
  \institution{The Chinese University of Hong Kong, Shenzhen}
  \country{Mainland, China}
}
\email{nanjunzheng@link.cuhk.edu.cn}

\author{Junhua Liu}
\affiliation{%
  \institution{The Chinese University of Hong Kong, Shenzhen}
  \country{Mainland, China}
}
\email{junhualiu@link.cuhk.edu.cn}

\author{Zhen Li}
\affiliation{%
  \institution{The Chinese University of Hong Kong, Shenzhen}
  \country{Mainland, China}
}
\email{zhenli@cuhk.edu.cn}


\begin{abstract}
We introduce a novel, multimodal large-scale scene reconstruction benchmark that utilizes newly developed 3D representation approaches: Gaussian Splatting and Neural Radiance Fields (NeRF). Our expansive U-Scene dataset surpasses any previously existing real large-scale outdoor LiDAR and image dataset in both area and point count. GauU-Scene encompasses over 6.5 square kilometers and features a comprehensive RGB dataset coupled with LiDAR ground truth. Additionally, we are the first to propose a LiDAR and image alignment method for a drone-based dataset. Our assessment of GauU-Scene includes a detailed analysis across various novel viewpoints, employing image-based metrics such as SSIM, LPIPS, and PSNR on NeRF and Gaussian Splatting based methods. This analysis reveals contradictory results when applying geometric-based metrics like Chamfer distance. The experimental results on our multimodal dataset highlight the unreliability of current image-based metrics and reveal significant drawbacks in geometric reconstruction using the current Gaussian Splatting-based method, further illustrating the necessity of our dataset for assessing geometry reconstruction tasks. We also provide detailed supplementary information on data collection protocols and make the dataset available on the following anonymous project \href{https://saliteta.github.io/anonymous_project/}{\color{red!50!white}{page}}.
\end{abstract}


\maketitle

\section{Introduction}
\begin{figure}
    \centering
    \includegraphics[width=0.48\textwidth]{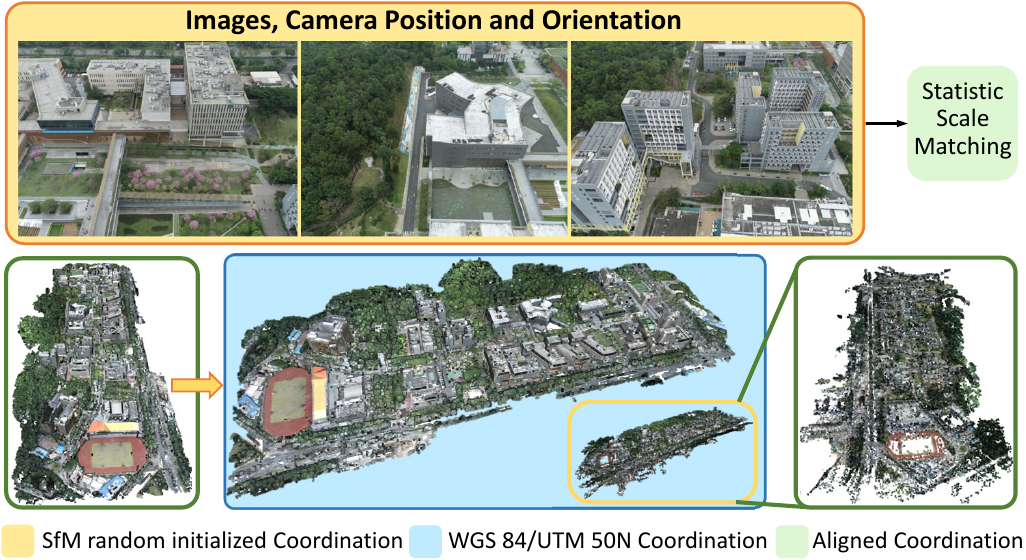}
    \caption{The dataset prepared for input into the neural field and Gaussian Splatting typically consists of camera positions and images in COLMAP format. The Structure from Motion (SfM) algorithm implemented in COLMAP initializes camera positions randomly, which may not align with LiDAR data in WGS 84 coordinates. This discrepancy poses a significant challenge for geometric alignment measurement and multi-modal fusion algorithms. When inputs are in two different coordinate systems, further validation becomes impractical. To address this, we propose a straightforward yet effective method for statistical scale matching to align LiDAR point clouds with camera positions. This approach is crucial for the construction of our dataset. The details of the preprocessing process will be introduced in Section~\ref{method}.}
    \label{fig: ProblemStatement}
\end{figure}

\begin{figure*}[htbp]
  \centering
    \includegraphics[width=0.95\textwidth]{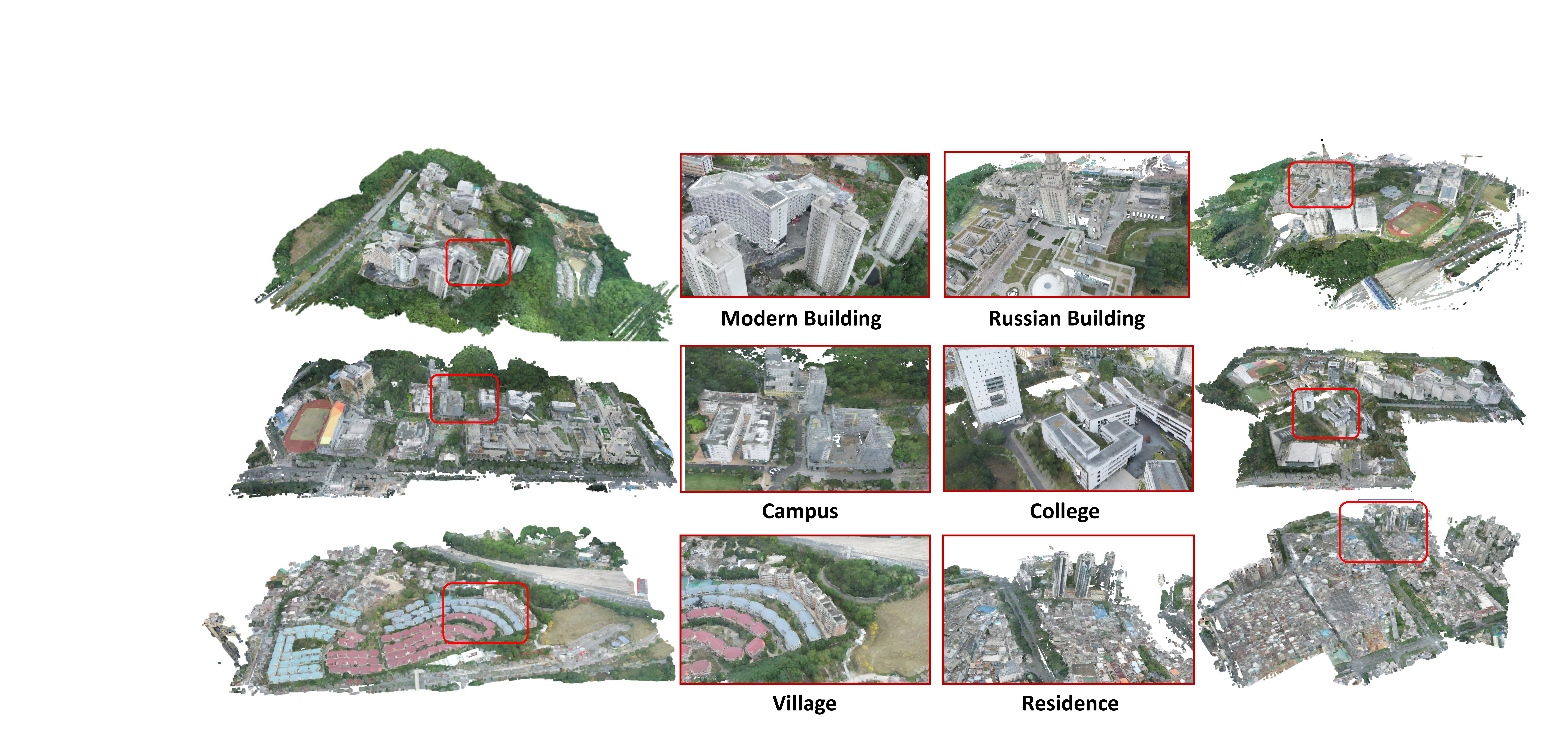}
    \vspace{-8pt}
   \caption{Our dataset is organized into six primary sections. The first and second scenes, located in the top row of the graph, feature the Modern Building and the Russian Building, respectively. The third and fourth scenes, depicted on the second line of the graph, represent a campus and a college. The last line combines a village and a residence. The dataset was collected using high-precision LiDAR and high-resolution cameras demonstrate its multimodal capabilities. The area it covers exceeds 6.5 km² and includes thousands of aligned images. Both the point cloud and images are aligned in the COLMAP coordinate system.}
   \label{fig:CUHKSZ_SMBU}
\end{figure*}

\begin{table*}[htbp]
\centering
\caption{This table provides detailed comparisons between our dataset and previously collected datasets. "Ptgy" stands for Photogrammetry, which is a non-LiDAR-based data acquisition method. Only real scenes are included in this table. We are the largest multimodal dataset in terms of lidar point number and area}
\vspace{-8pt}
\begin{tabular}{@{}lccccccc@{}}
\toprule
\textbf{Dataset}  & \textbf{Acquisition} & \textbf{Data Type} & \textbf{Area/Length} & \textbf{Image Number} & \textbf{Points/Triangular} & \textbf{scene}\\ \midrule
KITTI \cite{kitti}                    & Car Camera/Lidar & PC/Image      & 39.20$km$   & 300K    & 4549M            & 1 \\ 
BlockNeRF \cite{blocknerf}            & Car Camera       & Image         & -           & 12k     & -                & 1\\ 
MILL 19 \cite{megaNeRF}               & UAV Camera       & Image         & -           & 3.6k    & -                & 2\\ 
UrbanBIS\cite{UrbanBIS}               & UAV Ptgy         & PC/Mesh/Image & 10.78$km^2$ & 113.3k  & 2523.8M/284.3M   & 5\\ 
DublinCity \cite{dublincity}          & UAV Lidar        & PC/Image      & 2.00$km^2$  & -       & 260M             & 1\\ 
\midrule
Hessigheim \cite{kolle2021hessigheim} & UAV Camera/Lidar & PC/Mesh       & 0.19$km^2$ & -       & 125.7M/36.76M    & 1 \\
UrbanScene3D\cite{Urbanscene3D}       & UAV Camera/Lidar & PC/Image      & 3.03$km^2$  & \textbf{31k}   & 120M             & 6\\ 
GauU-Scene\cite{xiong2024gauu}       & UAV Camera/Lidar & PC/Image      & 1.5$km^2$  & 1.4k   & 210.6M             & 3\\ 
\midrule
\textbf{GauU-Scene V2}(Ours)                   & UAV Camera/Lidar & PC/Image      & \textbf{6.67}$km^2$ & 4.6k    & \textbf{627.5M}           & 6\\ 
\end{tabular}
\label{table:dataset_comparision}
\end{table*}
3D reconstruction is a transformative technology, enabling the conversion of real-world scenes into digital three-dimensional models. This technology, which often involves transforming multiple 2D images into 3D models, fosters applications in urban planning, virtual reality (VR), and augmented reality (AR). Various techniques have been employed to enhance the accuracy and efficiency of 3D reconstruction. Among them, Structure from Motion (SfM) has been notably prominent, as extensively studied by \cite{snavely2006photo}. This photogrammetric technique uses 2D images captured from different angles to reconstruct three-dimensional structures.

Recent innovations such as Neural Radiance Fields (NeRF) have shown great success in 3D reconstruction. NeRF predicts volumetric scene representations from sparse 2D images using a fully connected neural network \cite{mildenhall2021nerf}. Despite initial challenges such as training difficulty and limitations in scale and complexity, rapid advancements have been made. Techniques such as meta-learning \cite{chen2021mvsnerf}, sparsity exploitation \cite{Zhang_2022_CVPR}, data structure integration \cite{martel2021acorn, granskog2021neural}, and eigenvalue usage \cite{Chen2022ECCV} have significantly improved NeRF's performance. The current state-of-the-art method in addressing aliasing issues has also been proposed \cite{barron2022mip}. 3D Gaussian Splatting (3DGS) \cite{kerbl3Dgaussians} merges rasterization with novel view synthesis, features rapid training and rendering speeds, shows high tolerance to sparse camera positions and orientations, and offers an explicit representation that facilitates easy modification.

Adapting various deep representations to large-scale environments, such as cityscapes, has become an emerging research area. Modified Neural Radiance Field (NeRF) methods offer promising solutions for large-scale scene reconstruction tasks. For instance, \cite{blocknerf} opts to partition the entire scene into chunks and constructs the model separately for each. Meanwhile, \cite{xiangli2022bungeenerf} tackles the multi-scale challenge associated with the Google Earth dataset. Studies such as \cite{lin2024vastgaussian, citygs, GaussianPro} propose solutions based on 3DGS for large-scale scene reconstruction, utilizing the divide and conquer technique to partition the large-scale scene into blocks and render them one after another.

Efforts like \cite{scaffoldGaussian, SuGaR, 2DGS} aim to more closely align Gaussian Splatting with the underlying geometry. However, there is still a lack of a reliable dataset that is both sufficiently large and provides a reliable underlying geometry. We argue that current image-based metrics such as PSNR, LPIPS, and SSIM cannot adequately measure the underlying geometry of a 3D model, as demonstrated by our experiments, which further shows the value of our proposed accurate LiDAR point cloud.

The challenge of city-scale reconstruction extends beyond the model to include the dataset itself. Existing city-scale datasets, such as those mentioned in \cite{xiangli2022bungeenerf, UrbanBIS, Urbanscene3D, kitti, blocknerf}, have limitations, including the absence of accurate ground truth for point cloud or mesh data, or a focus on scene understanding rather than reconstruction. To overcome these challenges, we utilize the DJI Matrice 300 drone equipped with the Zenmuse L1 LiDAR to capture highly accurate 3D RGB point clouds as shown in Fig.~\ref{fig:CUHKSZ_SMBU}. This strategy allows us to explore beyond traditional indoor scenes and undertake extensive city-scale outdoor scene reconstructions. Our dataset spans an area larger than 6.5 km\(^2\), surpassing the scale of existing datasets such as UrbanBIS \cite{kolle2021hessigheim} and the UrbanScene3D real dataset \cite{Urbanscene3D} and \cite{xiong2024gauu}, which also used highly accurate LiDAR. Detailed comparisons are shown in Table~\ref{table:dataset_comparision}.

A primary challenge in drone-based datasets is their inability to simultaneously provide point cloud and image datasets, largely due to discrepancies in coordinate systems. In this work, we also propose a straightforward yet effective method to align Structure from Motion (SfM) camera positions with LiDAR data points.

Finally, we provide a detailed benchmark on the current popular off-the-shelf baselines including Vanilla 3D Gaussian Splatting~\cite{Gao_2023_SIGGRAPH}, SuGaR~\cite{SuGaR}, InstantNGP~\cite{InstantNGP}, and NeRFacto~\cite{nerfstudio}. The evaluation is performed on both image-based metrics, treating additional image data as the ground truth, and the Chamfer distance metric, using LiDAR point clouds as the ground truth.

To summarize, our threefold contributions to the field of 3D reconstruction and large-scale environmental mapping are as follows:

\begin{itemize}
    \item We introduce a comprehensive multimodal dataset captured using the DJI Matrix 300 drone equipped with the Zenmuse L1 LiDAR, providing highly accurate 3D RGB point clouds. Our dataset covers an area larger than 6.5 km\textsuperscript{2}, surpassing existing datasets in scale and detail, and is designed to support city-scale outdoor scene reconstructions.
    
    \item We propose a novel method for aligning Structure from Motion (SfM) camera positions with LiDAR data points, effectively overcoming the challenge of discrepancies in coordinate systems between point cloud and image datasets. This alignment technique enhances the integration of diverse data sources, crucial for accurate 3D reconstruction. And make the underlying geometry consistency measurement possible.
    
    \item We perform a detailed bench-marking of current popular 3D reconstruction methods, including SuGaR, Vanilla Gaussian Splatting, InstantNGP, and NeRFacto, providing valuable insights into their performance and applicability to large-scale reconstructions. This comparative analysis shows potential unreliability which further shows the vitality of our proposed Lidar point cloud.
\end{itemize}

In the following section, we will introduce related work, including Large Scale 3D Outdoor Datasets, Gaussian Splatting, and NeRF. We will then describe our dataset by illustrating its multimodal properties, scale, and data collection methods, as well as the LiDAR alignment method. Subsequently, we will provide a detailed experiment and explanation of the results, both qualitatively and quantitatively.

\begin{figure*}[htbp]
    \centering
    \includegraphics[width=\textwidth]{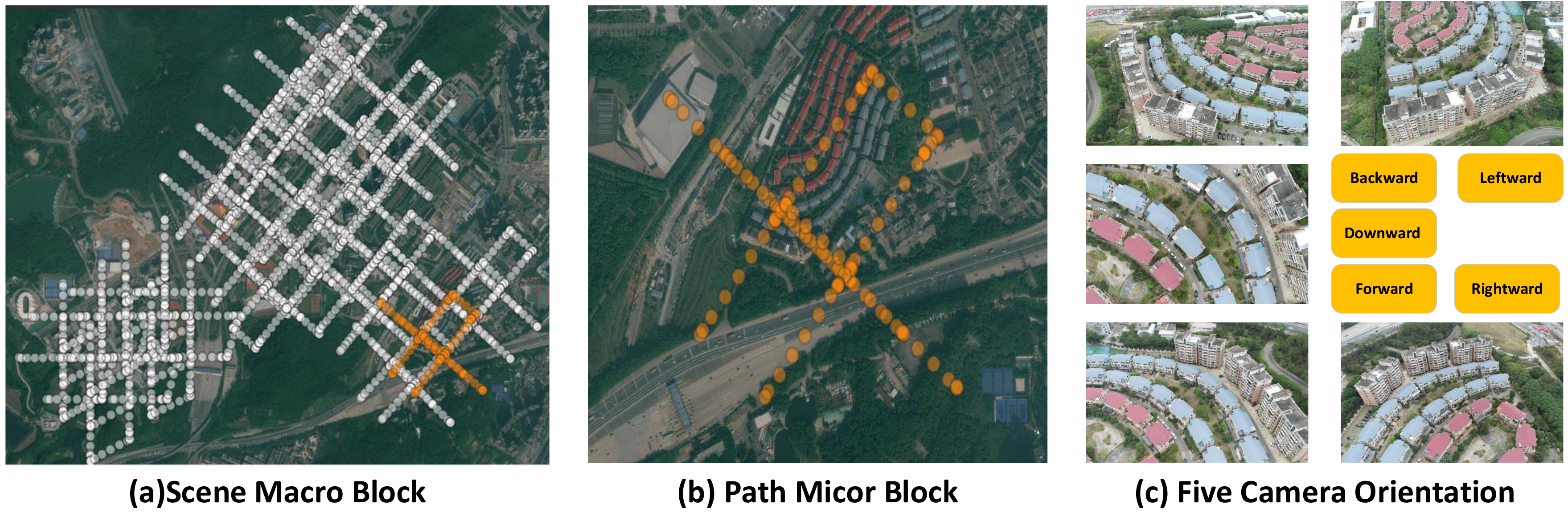}
    \caption{This figure shows the design of the drone routing path. The white and orange dots represent the positions where the drone took pictures. The overall path for a scene is shown in Graph (a), which is composed of several micro-blocks. One such micro-block, highlighted in orange, is detailed in Graph (a). Zooming into this orange micro-block reveals Figure (b). The total path length of each micro-block is limited by the battery life of the DJI Matrice 300, as well as the power consumption of the LiDAR in windy conditions. For safety reasons, each micro-block typically covers an area of $350\times350$ square meters. Each micro-block has five routing paths, providing different angles for photography, as illustrated in Figure (c). The first routing path offers a Bird's Eye View (BEV), while the subsequent four paths alter the camera's orientation by 45 degrees towards the horizontal plane. These four paths' camera orientations are forward, backward, rightward, and leftward, respectively.}
    \label{fig:routing}
\end{figure*}
\section{Related Work}

In subsequent sections, we discuss related work in 3D Large Scale Outdoor scene dataset followed by the related work in Gaussian Splatting and Nerual Radiance Field. 

\subsection{Large Scale 3D Outdoor Dataset}
Outdoor large-scale datasets, such as those mentioned in \cite{xiangli2022bungeenerf}, are captured using images of different scales without ground truth. Researchers typically evaluate their models by selecting novel views for generation and comparing them with the collected data. However, several images may not fully represent the entire 3D structure. In \cite{xiangli2022bungeenerf}, the use of varying scales from satellite-captured images introduces clear differences in time and a lack of 3D ground truth. For example, images captured in the morning and at night, or satellite images taken years apart (e.g., 2010 vs. 2020) of the same region, exhibit significant differences. We refer to this as the 'image time difference' problem in our work.

The KITTI dataset \cite{kitti} tries to provide ground truth point cloud data using LiDAR mounted on cars, but it falls short in capturing data from rooftops and higher buildings. The blockNeRF\cite{blocknerf} approach addresses the time difference through style transformation but does not offer a publicly available point cloud dataset in the Waymo Block-NeRF Dataset. \cite{UrbanBIS} labels the point cloud dataset collected in real cities, aiding in point cloud segmentation and understanding. Although the overall size of the dataset is over 10 $km^2$ but the data is constructed using multiview camera instead of high-precision Lidar. \cite{Urbanscene3D} employs drones with LiDAR to capture point cloud data, effectively overcoming the image time difference and rooftop data capture issues. However, most data from \cite{Urbanscene3D} are synthetic views generated by 3D rendering engines, lacking real-world complexity and the unclear correlation between LiDAR point clouds and images due to coordinate differences poses a significant challenge, limiting the use of either point cloud or image data from \cite{Urbanscene3D}. Other large-scale dataset such as\cite{mcdviral2024} 
 
In our current work, we tackle these challenges by providing a clear, straightforward pipeline to align datasets of different modalities. Moreover, our dataset is more than twice as large as those previously mentioned, offering a significant advancement in outdoor large-scale dataset research.
\begin{table*}[htbp]
\centering
\caption{
This table presents detailed coverage of scene reconstruction. We ensure that the size of each scene is maintained at approximately 1 km². This constraint limits the variation in lighting effects caused by the sun. The density of our point cloud is 20 cm per point. The raw data consists solely of DJI raw data and does not include the post-processed point cloud from the DJI Terra. The "Avg Height" denotes the average height of the drone's flight path relative to the altitude from which the drone took off. This height is consistently higher than that of the tallest local building. It is important to note that the maximum effective distance for LIDAR detection should be less than 250 m.}
\begin{tabular}{@{}lcccccc@{}}
\toprule
\textbf{Scene} & \textbf{Area in km\textsuperscript{2}} & \textbf{Image Number} & \textbf{Points Number} & \textbf{Raw Data in GB} & \textbf{Avg Height in $m$} & \textbf{Resolution}\\ \midrule
Campus & 1.020 & 670 & 79,767,884 & 12.5 & 120 & 5472 $\times$ 3648\\ 
Modern Building & 0.923 & 715 & 94,218,901 & 13.5 & 120 & 5472 $\times$ 3648\\ 
Village & 0.815 & 424 & 26,759,799 & 7.8 & 120 & 5472 $\times$ 3648\\ 
Residence & 1.467 & 1106 & 98,547,710 & 19.8 & 150 & 5472 $\times$ 3648\\ 
Russian Building & 0.908 & 563 & 283,31,405 & 16.2 & 150  & 5472 $\times$ 3648\\ 
College & 1.557 & 1215 & 58,979,628 & 22.3 & 136 & 5472 $\times$ 3648\\ 
\midrule
\textbf{Total} & 6.668 & 4693 & 627,500,327 & 92.1 & Nan & Nan\\ 
\end{tabular}

\label{table:dataset_details}
\end{table*}

\subsection{Gaussian Splatting}
Since the emergence of Gaussian Splatting \cite{kerbl3Dgaussians}, numerous studies have adopted 3D Gaussian Splatting (3DGS) as the primary representation for 3D models, including both human figures and environments. While some research has focused on identifying inherent issues with Gaussian Splatting, such as aliasing \cite{aliasing_3dgs}, many others have explored downstream applications. These applications span autonomous driving environments \cite{drivegaussian}\cite{yan2024street}\cite{gsSLAM} and human representation \cite{dreamGauZiwei}\cite{gauhumanZiwei}\cite{humanGauZiwei}\cite{GPSyebin}\cite{animatableyebin}\cite{headAvataryebin}.

In addition, there is significant literature on large-scale scene reconstruction. \cite{GaussianPro} introduces different methods by imposing additional constraints, including normals of Gaussian Splats and color initialization. VastGaussian \cite{lin2024vastgaussian} segments the entire 3D scene based on the locations of initialized points, camera positions, and other properties, subsequently merging the training results according to the initial points' locations. Although these studies offer reliable measurements using extra images as ground truth, the precise alignment of underlying geometry remains unproven. We plan to conduct experiments demonstrating that geometry alignment remains a substantial challenge in the Gaussian Splatting method and requires further improvement, as discussed in Section \ref{sec:EXP}.

Meanwhile, works such as \cite{scaffoldGaussian}\cite{SuGaR}\cite{2DGS} focus on enhancing the alignment of geometry. Scaffold Gaussian \cite{scaffoldGaussian} employs a neural network to generate Gaussian Splatting around anchor points, aiming to constrain the offset of Gaussian Splatting based on the viewing direction, potentially improving the alignment of 3D Gaussians. SuGaR \cite{SuGaR} introduces a regularization term that promotes the alignment of 3D Gaussians with the scene's surface and offers a method for mesh extraction by enforcing 3DGS to converge to a disk-like shape that is suitable for mesh extraction. Conversely, 2DGS \cite{2DGS} reduces 3D Gaussian Splats from an ellipsoid to a 2D disk directly, enabling a closer representation for geometric alignment. However, none of these studies validated their experimental results using a dataset with a 3D point cloud as ground truth. Thus, while they may be effective for novel view synthesis, their efficacy in 3D reconstruction settings remains unverified. Therefore, in the current study, we provide a detailed experiment on our dataset using SuGaR as test method.

The dataset introduced in this study addresses these issues by providing a reliable 3D point cloud that can serve as ground truth. This point cloud can be used as raw data or easily converted to a mesh for measuring the quality of 3D geometry alignment.

\subsection{Neural Radiance Field}
\cite{neuralField} and subsequent studies \cite{mildenhall2021nerf}\cite{Zhang_2022_CVPR}\cite{chen2021mvsnerf}\cite{martel2021acorn}\cite{granskog2021neural}, and\cite{barron2022mip} have demonstrated significant potential in novel view synthesis and scene reconstruction. However, their focus has predominantly been on indoor and bounded scene reconstructions. Attempts to extend these methods to large-scale scene reconstructions, such as by BlockNeRF\cite{blocknerf} and BunGeeNeRF\cite{xiangli2022bungeenerf}, face challenges, notably the absence of datasets and geometry extraction algorithms for comparison with real-world geometry. InstantNGP\cite{InstantNGP} and NeRFacto\cite{nerfstudio}, while not specifically designed for 3D outdoor scene reconstruction, offer efficient design and geometry extraction capabilities, making them suitable baselines for 3D reconstruction tasks. In our study, we will employ these algorithms to evaluate the Chamfer distance.

\begin{figure*}
    \centering
    \includegraphics[width = \textwidth]{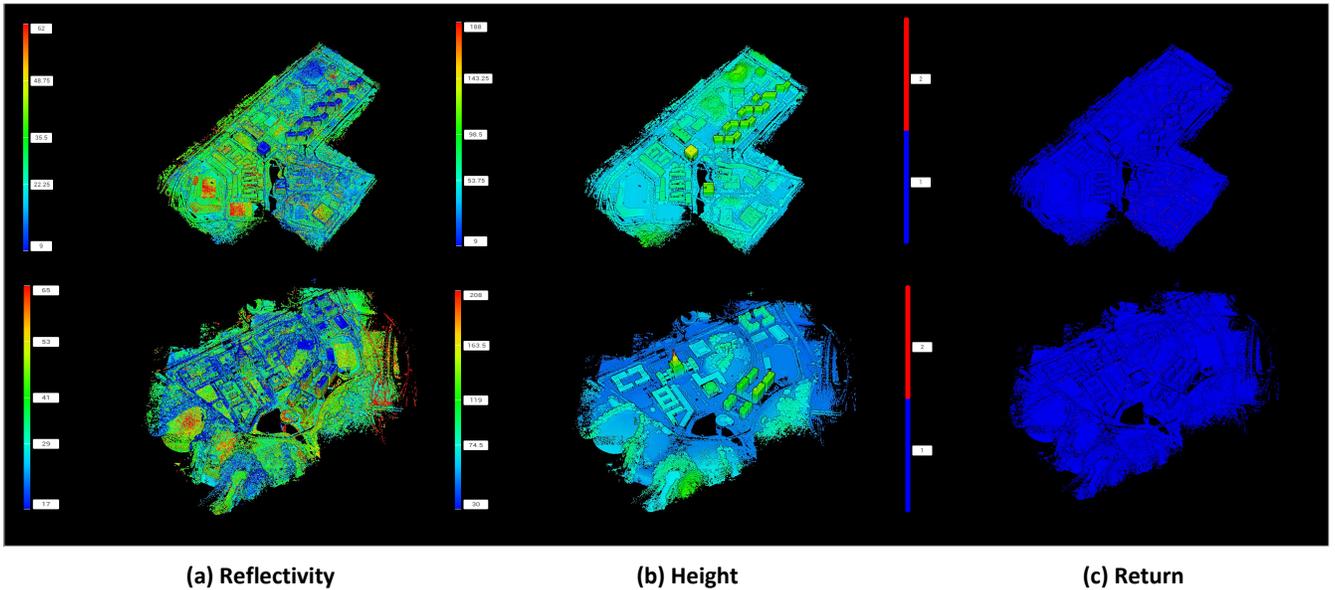}
    \vspace{-12pt}
    \caption{Our dataset provides essential information for quality control and multi-modal analysis and visualization. By using professional tools such as DJI Terra, one can observe three important properties critical for quality control: Reflectivity, Height, and Return. Graph (a) in this figure illustrates reflectivity, which measures the amount of light reflected back to the LiDAR sensor from surfaces or objects. Meanwhile, height, shown in graph (b), represents the building's altitude relative to the drone's takeoff altitude. The return, presented in graph (c), indicates the number of light returns detected by the LiDAR. Since our analysis filters out all data except those with at least two returns, moving objects, represented by red dots, will be excluded. More visualization results can be explored in our dataset or in the supplementary materials.}
    \label{fig:rhr}
\end{figure*}
\section{U Scene Dataset}
\label{method}
In this section, we will first introduce data property includes data format, data scale, and data collection methods in order. Then we will have a detailed  comparison and analysis of our dataset. 
\subsection{Data Property}
\subsubsection{Data Format}
We provide following information in our dataset:

\begin{itemize}[itemsep=0pt,parsep=1.5pt,leftmargin=*]
    \item Image COLMAP dataset with aligned Lidar point cloud
    \item Lidar Point Cloud in WGS 84/UTM 50N coordinates in ply format
    \item RAW PNTS points format used in DJI terra for quality control
    \item Point Reflectivity, height, and return
    \item Raw DJI routing logs in WGS 84/UTM 50N coordinates for camera position acquisition
    \item Raw DJI Lidar format 
\end{itemize}

\subsubsection{Data Scale}
We have six real-world scenes and corresponding LiDAR point clouds as provided in Fig.\ref{fig:CUHKSZ_SMBU}. The detailed report of the dataset is shown in Tab.\ref{table:dataset_details}. The dataset includes high-altitude buildings in various universities, residential areas, hospitals, vegetation, streets, and bridges. The altitudes range from -20 meters to 120 meters relative to the drone's take-off altitude. Due to the property of LiDAR device, we cannot capture the water and glasses.
 \begin{figure}[h]
    \centering
    \includegraphics[width=0.45\textwidth]{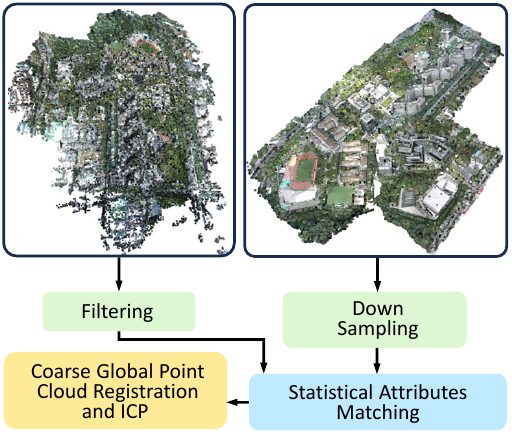}
    \caption{The orange color indicates the point cloud is in the COLMAP coordinate, and the blue color indicate the point cloud is in WGS coordinate. Before we use the proposed matching algorithms, we need to filter the COLMAP point cloud according to the points' distance to mean, and down sampling the Lidar point cloud. Then resealing lidar point cloud and manually registration or ICP registration}
    \label{fig:alignment}
\end{figure}
 
\subsubsection{Data Collection Method}
 In this section we will introduce the data collection method post-processing method, and coordinates alignment method in detail.
 
\textbf{Data Collection Method:}
\textit{Drone Assembly and Hardware Preparation:} We use the Matrix300 as the carrying platform, equipped with the Zenmuse L1 Lidar for data acquisition. Details on drone assembly and hardware management can be found in the supplementary. We provide accurate description and image guidance. 

\textit{Path Planning:} Our dataset provides images with multiple different camera orientations. As shown in Fig.\ref{fig:routing}, we employ oblique shooting in DJI Terra for path planning. We set the oblique angle to 45 degrees, we control the scanning space of one route suitable for the battery life. The detailed inforamtion can be found in supplementary. 

\textit{Post-Porcessing: } We use DJI Terra, an industrialized drone collected data postprocessing software that easily transfer the raw lidar and image data to PLY format data in WGS 84/UTM 50N coordinates and report detailed information for point cloud quality. As shown in Fig.\ref{fig:rhr} and Fig.\ref{fig:CUHKSZ_SMBU}. We also provide a detailed guidance in the supplementary, However, current DJI Terra has gone through several major changes, therefore, our guidance might not suitable for the newest version.

\textit{Wind and Vibrant Prevention: } We will not take off when the speed of wind is larger than 10 meters per second. Therefore, the vibration brought by the wind is controlled and therefore further improves the accuracy of our dataset.

\textbf{Quality Control:}
According to the specifications of the Lidar and the drone's flying platform from DJI official website, we ensure the flying height is always less than 150 meters. The raw data points, without any filtering, will have a standard deviation of less than 3 cm. By utilizing DJI Terra, an industrial solution for Lidar data post-processing software, we can limit the error to less than 5 cm when the real distance from the detected object to the Lidar is less than 300m. Our data acquisition time is constrained during 2:00pm to 5:00pm in the afternoon so that the lighting effect does not change significantly. By utilizing double return technique, we remove all object that is moving in the scene and it improves the point cloud accuracy at the same time. 

\textbf{Transformation Matrix obtained:} To address the LiDAR-image alignment problem we mentioned above, we propose a simple pipeline that is essential for constructing a usable multimodal dataset. We have observed that most methods used in scene reconstruction utilize COLMAP \cite{colmap1}\cite{colmap2} to determine camera orientation. A byproduct of this procedure is a sparse point cloud. Transferring the camera orientation to WGS coordinates is costly, as DJI does not provide such information. Therefore, instead of transferring COLMAP to WGS coordinates, we register our dense LiDAR point cloud to the sparse COLMAP point cloud. We first voxel downsample the LiDAR point cloud to a reasonable density. Since the COLMAP and LiDAR point clouds differ in translation, scale, and orientation, we begin by aligning them through a simple scaling.
\begin{equation}
    T_l = \mu_c - \mu_l, S_l = \frac{\sigma_c}{\sigma_l} \\
    \hat{P_l} = (P_l - T_l)S_l
\end{equation}

Given Lidar point cloud ${p_l} = \{p_{l1}, \ldots,p_{lM}\}$ and COLMAP point cloud $P_c = \{P_{c1}, \ldots, P_{cN}\}$ in $\mathbb{R}^3$. We first obtain scaled Lidar Point Cloud $\hat{p_l}$ utilizing a simple scaling shown in the above equation. Where $\mu_c$ represent the mean point of COLMAP sparse point cloud, $\mu_l$ represents the mean of down sampled Lidar point cloud. $\sigma_c,\sigma_l$ represents the standard deviation of COLMAP and Lidar point cloud respectively. After this step we might need to rotate the point cloud manually or simply apply ICP. We optimize a rigid transformation on $\hat{P_l}$ (represented using a rotation matrix $R \in \mathbb{R}^{d \times d}$ and a translation vector $t \in \mathbb{R}^d$) to align $P_l$ with $P_c$:
\begin{equation}
\min_{R,t} \sum_{i=1}^{M} (D_i(R,t))^2 + I_{SO(d)}(R),
\end{equation}
where $D_i(R,t) = \min_{q \in Q} \|R\hat{p_l}_i + t - p_c\|$ is the distance from the transformed point $Rp_i + t$ to the target set $P_c$, and $I_{SO(d)}(\cdot)$ is an indicator function for the special orthogonal group $SO(d)$, which requires $R$ to be a rotation matrix:

\begin{equation}
I_{SO(d)}(R) = 
\begin{cases} 
0, & \text{if } R^TR = I \text{ and } \det(R) = 1,\\
+\infty, & \text{otherwise}.
\end{cases}
\end{equation}

After obtaining the transformation matrix and scalling factor, we can easily transfer the Lidar Point cloud to COLMAP point cloud. 

\subsection{Analysis and Comparison}
Our dataset offers several advantages compared to other datasets. Firstly, we utilize the Zenmuse L1 to obtain ground truth geometry, whereas most datasets \cite{blocknerf}\cite{megaNeRF}\cite{UrbanBIS} rely on monocular or multi-view cameras for data acquisition, which are more suitable for novel view synthesis than for scene reconstruction. Our dataset provides city-scale information, including tall buildings, lakes, mountains, and rooftops, features that are less commonly provided by others \cite{kitti}\cite{blocknerf}. It is more than twice as large as \cite{Urbanscene3D} in terms of area and offers significantly larger point cloud sizes compared to \cite{kolle2021hessigheim}\cite{Urbanscene3D}.

We employ a double-return technique to remove all foreground or moving objects, ensuring a stable lighting effect. Additionally, we maintain consistent partitioning of different scenes, which is more suitable for reconstructed scene loading in Gaussian Splatting and NeRF. The details of this comparison are shown in Table \ref{table:dataset_comparision}. Our dataset stands out among those providing highly accurate LiDAR data, featuring a much larger scale in both area and point number. Although we provide fewer images than previous methods, the information in our images is comparable, as we remove redundant information between consecutive images during the flight route. We provide detailed transformation matrix to combine both Lidar information as both ground truth and for future multi modal learning. Furthermore, we utilize Lidar point cloud in our following experiment, the results yield potential gap in current image based measurment matrix.

\begin{table*}[htbp]
\centering
\label{Tab:method_comparison_1}
\caption{This table displays the results obtained when testing our dataset with different methods, including SuGaR\cite{SuGaR} and NeRFacto\cite{nerfstudio} We measured the training time in terms of GPU count multiplied by training time in minutes. For training and evaluating the Gaussian Splatting results, we used the official implementation of SuGaR with there evaluation code. While for the geometric measurement, we use the refined point cloud as the final product for comparison. Meanwhile, the NeRF Studio implementation was utilized for NeRFacto to conduct training and evaluation. }
\label{tab:real_q}
\resizebox{1\linewidth}{!}{
\begin{tabular}{l|cccccc|cccccc}
\toprule
Method & \multicolumn{6}{c|}{SuGaR} & \multicolumn{6}{c}{NeRFacto} \\
\begin{tabular}{c|c} Scene & Metrics \end{tabular}  & PSNR \(\uparrow\) & SSIM \(\downarrow\) & LPIPS \(\downarrow\) & Time~(GPU$\cdot$min) & Mean Err \(\downarrow\) & Std Err \(\downarrow\) & PSNR \(\uparrow\) & SSIM \(\uparrow\) & LPIPS \(\downarrow\) & Time~(GPU $\cdot$ min) & Mean Err \(\downarrow\) & Std Err \(\downarrow\)\\
\midrule

Campus          & 23.02 & 0.601 & 0.506 & 104 & 0.141 & 0.261 & 17.70 & 0.455 & 0.779 & 1692 & 0.067 & 0.198 \\

Modern Building       & 22.51 & 0.572 & 0.497 & 108 & 0.176 & 0.534 & 18.66 & 0.448 & 0.734 & 1704 & 0.050 & 0.170 \\

Village               & 22.78 & 0.619 & 0.461 & 98 & 0.160 & 0.304 & 16.95 & 0.399 & 0.727 & 1788 & 0.065 & 0.205 \\

Residence             & 20.97 & 0.533 & 0.607 & 119 & 0.167 & 0.638 & 15.05 & 0.364 & 0.879 & 1780 & 0.277 & 0.245 \\

Russian Building      & 21.58 & 0.618 & 0.450 & 103  & 0.338 & 0.599 & 16.61 & 0.405 & 0.682 & 1716 & 0.066 & 0.240 \\

College               & 22.02 & 0.588 & 0.514 & 123 & 0.154 & 0.194 & 17.28 & 0.462 & 0.781 & 1732 & 0.034 & 0.110 \\

\hline

Avg                   & 22.14 & 0.589 & 0.506 & 109.2 & 0.189 & 0.422 & 17.04 &0.422 & 0.764 & 1735.3 & \textbf{0.093} & \textbf{0.194} \\

\bottomrule

\end{tabular}}
\end{table*}

\begin{table*}[htbp]
\centering
\caption{This table displays the results obtained when testing our dataset with vanilla Gaussian Splatting and InstantNGP,  For training and evaluating the Gaussian Splatting results, we used the official implementation of Gaussian Splatting. Meanwhile, the NeRF Studio implementation was utilized for Instant-NGP to conduct training and evaluation.}
\label{tab:real_q}
\resizebox{1\linewidth}{!}{
\begin{tabular}{l|cccccc|cccccc}
\toprule
Method & \multicolumn{6}{c|}{Gaussian Splatting} & \multicolumn{6}{c}{Instant NGP}  \\
\begin{tabular}{c|c} Scene & Metrics \end{tabular}  & PSNR \(\uparrow\) & SSIM \(\downarrow\) & LPIPS \(\downarrow\) & Time~(GPU$\cdot$min) & Mean Err \(\downarrow\) & Std Err \(\downarrow\) & PSNR \(\uparrow\) & SSIM \(\uparrow\) & LPIPS \(\downarrow\) & Time~(GPU $\cdot$ min) & Mean Err \(\downarrow\) & Std Err \(\downarrow\)\\
\midrule

Campus & 24.76 & 0.735 & 0.343 & 58  & 0.079 & 0.207 
             & 20.76 & 0.516 & 0.817 & 220 & 0.123 & 0.378 \\

Modern Building &  25.49 & 0.762 & 0.273 & 64  & 0.096 & 0.312
             &  20.25 & 0.522 & 0.816 & 392 & 0.082 & 0.260 \\
                
Village & 26.14 & 0.805 & 0.237 & 62  & 0.124 & 0.305 
        & 20.79 & 0.511 & 0.792 & 268 & 0.177 & 0.497  \\

Residence & 22.03 & 0.678 & 0.371 & 71  & 0.248 & 0.192 
            & 18.64 & 0.453 & 0.856 & 348 & 0.228 & 0.314 \\
    
Russian Building & 23.90 & 0.784 & 0.248 & 63 & 0.186 & 0.440  
               & 18.37 & 0.507 & 0.810 & 252& 0.153 & 0.458 \\

College  & 24.21 & 0.749 & 0.326 & 64  & 0.064 & 0.168 
                    & 19.64 & 0.551 & 0.820 & 276 & 0.136 & 0.438 \\

\hline

Avg & \textbf{24.42} & \textbf{0.752} & \textbf{0.300} & \textbf{63.7} & 0.133 & 0.271 & 19.74 & 0.510 & 0.815 & 292.7 & 0.149 & 0.391 \\

\bottomrule
\end{tabular}}

\label{Tab:method_comparison_2}
\end{table*}

\section{Experiment and Result}
\label{sec:EXP}
In this section, we present the results of testing various baseline methods on our extensive dataset. We employed three different methods: Vanilla Gaussian Splatting \cite{kerbl3Dgaussians}, InstantNGP \cite{InstantNGP}, and NeRFacto \cite{nerfstudio}. Beyond the conventional image-based evaluation metrics such as PSNR, SSIM, and LPIPS, we also provide a reliable measure of geometric scene reconstruction accuracy using the Chamfer distance. To generate 3D point clouds for InstantNGP and NeRFacto, we used the export method available in NeRFStudio; for Vanilla Gaussian Splatting, we utilized the mean splatting point to form the point cloud.

We trained our model for Vanilla Gaussian Splatting using the official implementation, and for InstantNGP and NeRFacto, we utilized the NeRFStudio implementation. The Vanilla Gaussian Splatting model was trained on a single RTX 3090 GPU, while the NeRF-based models utilized four RTX 3090 GPUs. We assessed our image-based performance using a test dataset randomly selected to include 10\% of the images from each scene, with the remaining 90\% used for training. The detailed comparison of the methods is shown in Tables \ref{Tab:method_comparison_1} and \ref{Tab:method_comparison_2}. As can be clearly seen, 3DGS (assumed to be Vanilla Gaussian Splatting) and SuGaR demonstrate superior performance in image-based rendering scores, including PSNR, SSIM, and LPIPS, compared to the NeRF-based methods, while also achieving significantly shorter training times. This measurement result aligns with our expectations; more visualized results can be checked on our project page. SuGaR, as expected, shows less performance on image-based metrics since it sacrifices its ability to fit the lighting effects to form a better underlying geometry.

We further provide detailed comparisons by utilizing our highly accurate point cloud as ground truth. The Chamfer distance, embedded in CloudCompare, serves as the metric to measure discrepancies between two sets of point clouds. We use the mean distance and the standard deviation of the distance to represent how well a method performs on our dataset. The smaller the mean distance, the better the geometric reconstruction performance. For NeRF-based methods, we employ ns-export to generate 3D point clouds and use the standard method from Open3D for point cloud reconstruction. The mean value of every Gaussian Splatting instance is used as the exported point cloud for Gaussian Splatting. Point clouds generated by neural radiance fields for unbounded scenes often contain many outliers unrelated to the scene itself, leading to lower scores. Similarly, 3DGS suffers from edge effects, where the edges of each scene introduce significant blurring. As for SuGaR, it generates outliers that go beyond our expectations, which might be the reason it obtains a lower score in geometric reconstruction.
\begin{figure*}[h]
    \centering
    \includegraphics[width=\linewidth]{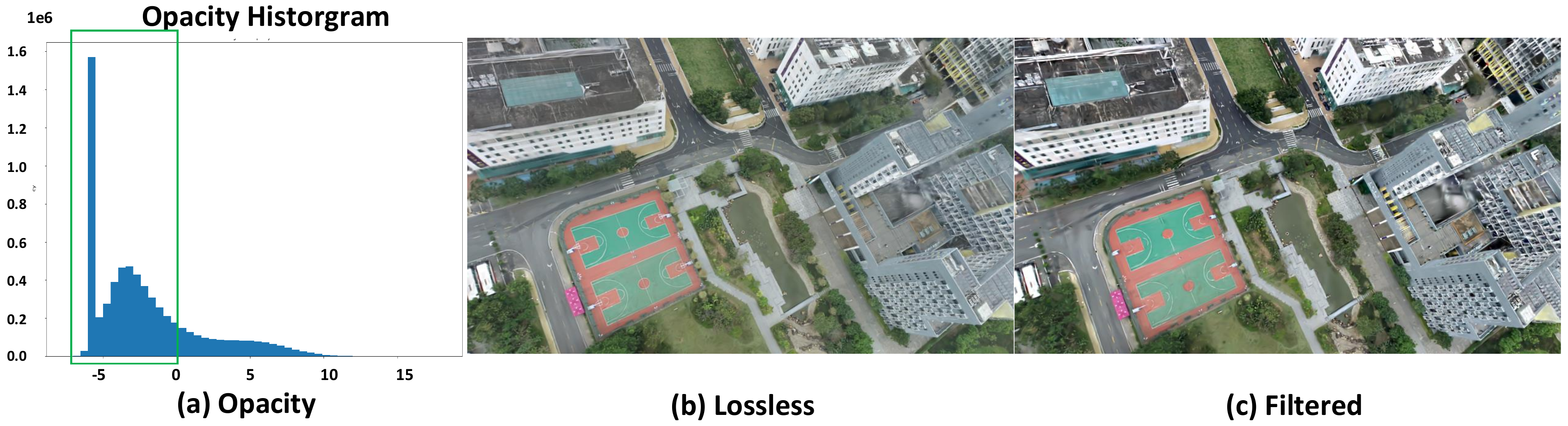}
    \vspace{-10pt}
    \caption{Figure (a) shows the statistics of 3DGS's logits for the alpha value, i.e., after passing this value through a sigmoid function, we obtain the final alpha value, which is constrained between 0 and 1. The original Gaussian Reconstruction is depicted in Figure (b). After removing all the Gaussian Splats encircled with green bounding boxes from (a), we obtain the filtered result, shown in Figure (c). The 3DGS we removed accounts for two-thirds of all 3DGS.}
    \label{fig:transparent}
\end{figure*}
A detailed comparison can be found in Table \ref{Point_difference}. Qualitative results are available on our project page. The LiDAR dataset we use will be downsampled to one-hundredth of its original size for simplification of calculations. Further detailed comparisons depend on downstream tasks. As one can easily find, NeRFacto, which obtained the lowest score in the image-based metrics, has the shortest Chamfer distance, while Instant-NGP and SuGaR are ranked at the last and second last places, respectively. This experimental result reveals the essential fact that image-based measurements cannot represent the underlying geometry, which further validates the importance of our current work. 

\begin{figure}[h]
    \centering
    \includegraphics[width=\linewidth]{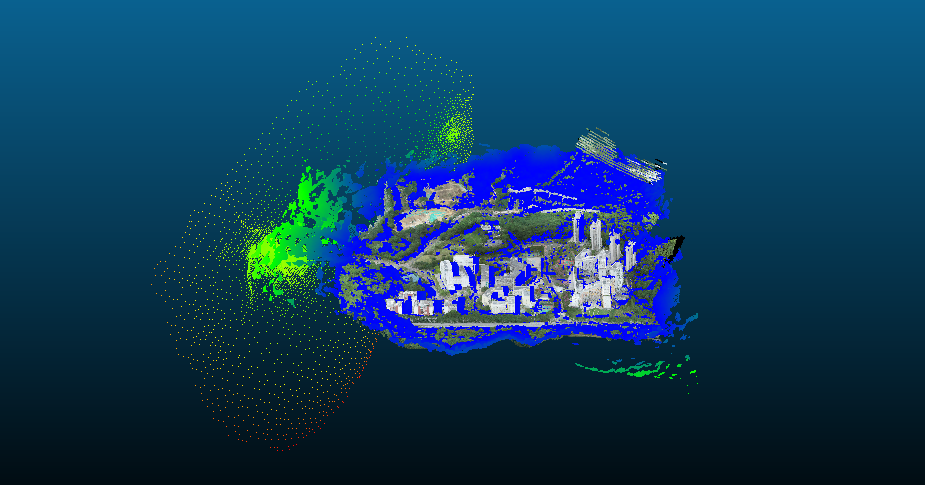}
    \vspace{-10pt}
    \caption{This figure demonstrates that SuGaR generates a mesh with outliers. The points in true color perfectly match the ground truth, while other points range from cold to warm colors, indicating outliers ranked by their distance. Other methods do not suffer from this high number of outliers. We argue that these outliers are the primary reason why SuGaR does not exhibit better performance in geometric construction results.}
    \label{fig:mesh}
\end{figure}

One might argue that using the mean of 3DGS to represent the corresponding points may not be the best approach. However, even if developing a better method could lead to 3DGS being ranked first in terms of geometric-based measurement metrics, InstantNGP should be ranked second, not last. Additionally, one might argue that SuGaR is a method designed for Geometric Alignment, yet it ranks last in terms of Chamfer distance. We further conducted a qualitative comparison and found that SuGaR indeed has better performance in geometric reconstruction. However, it generates a mesh with outliers, as shown in Fig.\ref{fig:mesh}. While blue points in the graph are common in other methods, green and even slightly red points are rarely seen in other methods, let alone in such large numbers. Detailed quantitative results will be provided in the supplementary materials. From a quantitative perspective, SuGaR indeed is the best method if we ignore the outliers shown here.

We further provide a potential qualitative analysis of image reconstruction that could explain why vanilla Gaussian Splatting fails to accurately represent the underlying geometry. In our simple analysis of the alpha values of Gaussian Splatting, we were surprised to find that almost two-thirds of them are nearly transparent, indicating very small alpha values. By removing these Gaussian Splats, we observed that the rendered image becomes sharper with virtually no loss of information, as illustrated in Fig.\ref{fig:transparent}. We argue that these nearly transparent instances of Gaussian Splatting float around in 3D space. Although they are not visible in the rendered image, they contribute to the degradation of geometric measurement metrics. One potential reason for the floating Gaussian is changes in weather. For example, in some overlapping regions of a picture, one shot may be taken when there is a cloud overhead, while another may be taken when the sky is clear. To optimize for image-based loss, the Gaussian Splats may tend to move away from the object surface towards the camera, appearing as purely black or white semi-transparent colors.

\section{Conclusion}
We present a large-scale dataset, Uscene, which utilizes both highly accurate Lidar and cameras to obtain multimodal information, including point clouds and images. We proposed a simple yet effective data preprocessing pipeline that can easily align Lidar point clouds and SfM image orientations together, which is vital for constructing our dataset. We further provide a detailed experimental analysis of our dataset using both images and point clouds as ground truth. The difference between experiments using images as ground truth and point clouds as ground truth reveals a potential gap for current image-based measurement metrics, including SSIM, LPIPS, and PSNR. This further indicates that the current 3DGS method has drawbacks in underlying geometry reconstruction.

However, the current geometry-based metric is not reliable either. As we mentioned in the experimental results, using the mean of 3DGS as the corresponding point is not dependable. In the future, we might develop more complex methods for geometry extraction. Moreover, apart from point clouds, meshes are also a traditional representation of 3D objects. We need to develop corresponding algorithms to measure the differences between Gaussian Splatting and meshes. SuGaR and its subsequent work might suffer from the outlier mesh, as we mentioned before, but the reason why SuGaR leads to unpredictable mesh generation remains unclear. The semi-transparent Gaussian Splatting generated by different lighting effects also presents a challenging issue for refining input images or future training methods.

\section{Acknowledgments}
This work is support by The Future Network of Intelligence Institute (FNii) at the Chinese University of Hong Kong, Shenzhen. The institute is undertaking relevant National Key R\&D Project, Guangdong Key R\&D project, and has set up the CUHK(SZ)-SRTC Spectrum Big Data Joint Lab, the CUHK(SZ)-JD Joint AI Lab, and Shenzhen key laboratory of Big Data and Artificial Intelligence. 

And Butian Xiong is supported by URA(Undergraduate Research Award) by The Chinese University of Hong Kong, Shenzhen from 2024 Jan to 2024 March

\newpage
\bibliographystyle{ACM-Reference-Format}
\bibliography{sample-base}

\end{document}